\documentclass[conference]{IEEEtran}
\IEEEoverridecommandlockouts
\usepackage{cite}
\usepackage{amsmath,amssymb,amsfonts}
\usepackage{algorithmic}
\usepackage{graphicx}
\usepackage{textcomp}
\usepackage{xcolor}
\def\BibTeX{{\rm B\kern-.05em{\sc i\kern-.025em b}\kern-.08em
    T\kern-.1667em\lower.7ex\hbox{E}\kern-.125emX}}

\usepackage{fancyhdr}
\usepackage{censor}
\usepackage{hyperref}
\usepackage{cleveref}
\usepackage{tikz}
\usepackage{standalone}
\usetikzlibrary{calc}
\usepackage{tabularx}
\usepackage{booktabs}
\usepackage{svg}
\usepackage{subcaption}
\usepackage{float}
\usepackage{orcidlink}

\usepackage[nolist,nohyperlinks]{acronym}

\begin{acronym}
  \acro{fer}[FER]{Facial Expression Recognition}

  \acro{jepa}[JEPA]{Joint-Embedding Predictive Architecture}
  \acroplural{jepa}[JEPAs]{Joint-Embedding Predictive Architectures}

  \acro{vjepa}[V-JEPA]{Video Joint-Embedding Predictive Architecture}
  \acroplural{vjepa}[V-JEPAs]{Video Joint-Embedding Predictive Architectures}

  \acro{ravdess}[RAVDESS]{Ryerson Audio-Visual Database of Emotional Speech and Song}

  \acro{cremad}[CREMA-D]{Crowd-sourced Emotional Multimodal Actors Dataset}

  \acro{hci}[HCI]{Human Computer Interaction}

  \acro{ac}[AC]{Affective Computing}

  \acro{maedfer}[MAE-DFER]{Efficient Masked Autoencoder for Self-supervised Dynamic Facial Expression Recognition}

  \acro{mae}[MAE]{Masked Autoencoder}
  \acroplural{mae}[MAEs]{Masked Autoencoders}

  \acro{dfer}[DFER]{Dynamic Facial Expression Recognition}

  \acro{videomae}[VideoMAE]{Video Masked Autoencoder}
  \acroplural{videomae}[VideoMAEs]{Video Masked Autoencoders}

  \acro{vit}[ViT]{Vision Transformer}

  \acro{uar}[UAR]{Unweighted Average Recall}

  \acro{war}[WAR]{Weighted Average Recall}

  \acro{mlp}[MLP]{Multi-Layer Perceptron}

\end{acronym}

\makeatletter
\newcommand{\linebreakand}{%
  \end{@IEEEauthorhalign}
  \hfill\mbox{}\par
  \mbox{}\hfill\begin{@IEEEauthorhalign}
}
\makeatother

\begin{document}

\title{Video Joint-Embedding Predictive Architectures for Facial Expression Recognition\\
\thanks{This work was funded in parts by the BIGEKO project (BMBF, German Ministry for Education and Research, grant number 16SV90904), TherapAI project (DFG, German Research Foundation, grant number 49316921), and FORSocialRobots project (BFS, Bavarian Research Foundation, grant number AZ1594-23).}
}

\author{\IEEEauthorblockN{1\textsuperscript{st} Lennart Eing}
\IEEEauthorblockA{\textit{Chair for Human-Centered Artificial Intelligence} \\
\textit{University of Augsburg)}\\
Augsburg, Germany \\
\href{mailto:lennart.eing@uni-a.de}{lennart.eing@uni-a.de}\orcidlink{0009-0005-8855-7659}}
\and
\IEEEauthorblockN{2\textsuperscript{nd} Cristina Luna-Jiménez}
\IEEEauthorblockA{\textit{Chair for Human-Centered Artificial Intelligence} \\
\textit{University of Augsburg}\\
Augsburg, Germany \\
\href{mailto:cristina.luna.jimenez@uni-a.de}{cristina.luna.jimenez@uni-a.de}\orcidlink{0000-0001-5369-856X}}
\linebreakand
\IEEEauthorblockN{3\textsuperscript{rd} Silvan Mertes}
\IEEEauthorblockA{\textit{Faculty of Computer Science} \\
\textit{Technical University of Applied Sciences Augsburg}\\
Augsburg, Germany \\
\href{mailto:silvan.mertes@tha.de}{silvan.mertes@tha.de}\orcidlink{0000-0001-5230-5218}}
\and
\IEEEauthorblockN{4\textsuperscript{th} Elisabeth André}
\IEEEauthorblockA{\textit{Chair for Human-Centered Artificial Intelligence} \\
\textit{University of Augsburg}\\
Augsburg, Germany \\
\href{mailto:elisabeth.andre@uni-a.de}{elisabeth.andre@uni-a.de}\orcidlink{0000-0002-2367-162X}}
}

\maketitle
\thispagestyle{fancy}

\begin{abstract}
    This paper introduces a novel application of Video Joint-Embedding Predictive Architectures (V-JEPAs) for Facial Expression Recognition (FER). 
    Departing from conventional pre-training methods for video understanding that rely on pixel-level reconstructions, V-JEPAs learn by predicting embeddings of masked regions from the embeddings of unmasked regions.
    This enables the trained encoder to not capture irrelevant information about a given video like the color of a region of pixels in the background.
    Using a pre-trained V-JEPA video encoder, we train shallow classifiers using the RAVDESS and CREMA-D datasets, achieving state-of-the-art performance on RAVDESS and outperforming all other vision-based methods on CREMA-D (+1.48 WAR). 
    Furthermore, cross-dataset evaluations reveal strong generalization capabilities, demonstrating the potential of purely embedding-based pre-training approaches to advance FER.
    We release our code at \href{https://github.com/lennarteingunia/vjepa-for-fer}{https://github.com/lennarteingunia/vjepa-for-fer}.
\end{abstract}


\section{Introduction}\label{sec:introduction}\noindent
    Facial expressions are an important means of human communication~\cite{kaulard2012mpi}.
    A large part of human emotion is communicated through the face, body posture, and other visual cues.
    For a long time, people have been trying to find common ground on which to describe and classify them.
    Notable instances of this are the theories of discrete universal emotions by Ekman~\cite{ekman1969repertoire} or the bidimensional valence-arousal theory of Russel~\cite{russell1980circumplex}.
    Emotion Recognition --- the ability to identify and interpret human emotions within these theories --- has emerged as a crucial field in both artificial intelligence and human-computer interaction research.
    As technology continues to evolve, understanding emotions has become essential not only for improving user experience, but also for improving the functionality of systems ranging from virtual assistants and customer service bots to healthcare monitoring systems and autonomous vehicles~\cite{sajjad2023comprehensive}.
    Emotion Recognition is still an active research field due to its intrinsic subjective nature and the complexity of categorizing emotions.
    
    In recent years, methods for automatic Emotion Recognition have largely shifted towards deep learning approaches, which require large-scale training datasets.
    However, producing and labeling \ac{fer} datasets is time-consuming and expensive.
    The resulting data scarcity makes it difficult to reliably train and evaluate any \ac{fer} models.
    Problems of data scarcity are not specific to the \ac{fer} problem however, but also arise in many other areas of applied computer vision.
    
    Methods for self-supervised pre-training have shown to be effective in a great number of computer vision applications.
    These methods promise to learn representations of data that can be easily used for any downstream task from more easily available unlabeled data.
    Most of these methods learn representations of their input data by applying perturbations to it, like masking regions of the input image or video and then reconstructing the perturbed inputs.
    A good example of this are methods based on \acp{mae}~\cite{he2022masked} for images and its recent extension \acs{videomae}~\cite{tong2022videomae}.
    Whether pixel-level reconstructions are a good pretext task is an ongoing discussion in the representation learning community.
    The problem lies in the fact that performing pixel-level reconstructions requires signal embeddings to retain pixel-level information about the input signal, which might be irrelevant to a given downstream task.
    This pixel-level information could include information about the color of a small region of pixels in the background for example, information that is not relevant when we want to identify changes in perceived emotion in a given video of a human face.
    Although most modern, well-performing vision encoder models are overparameterized, pixel-level information can sometimes pollute their representations with unnecessary detail.
    We hypothesize that this type of pixel-level information and therefore reconstruction pretext tasks are not necessary for \ac{fer}.
    
    In this work we show that by employing a \ac{vjepa}~\cite{bardes2024revisitingfeaturepredictionlearning} as the video encoder in a \ac{fer} setting, we are able to achieve state-of-the-art results without any additional required fine-tuning of the underlying video encoder.
    \acp{vjepa} are not trained on pixel-level reconstructions during their pre-training.
    They are instead trained to directly predict the embeddings of masked regions of video from the embeddings of the unmasked regions of video.
    Using an off-the-shelf model, we freeze all its weights and train a set of shallow classifiers using two lab-controlled \ac{fer} datasets.
    Then, we compare our results to those achieved using models pre-trained on pixel-level pretext reconstruction tasks, specifically developed for the \ac{fer} task.
    Our classifiers show state-of-the-art performance on both datasets, in some cases also generalizing well during cross-dataset evaluation.
    Our main contributions are:
    \begin{enumerate}
      \item We show that, given large-scale task-agnostic pre-training, it is possible to reach state-of-the-art \ac{fer} performance without any pixel-level pretext tasks and finetuning of the underlying video encoder model. 
      \item To our knowledge, we are the first to apply \ac{vjepa}, or any \ac{jepa} for that matter, to the \ac{fer} task.
      While there are other methods for recognition of affect-related states, e.g~\cite{stiber2024uh}, we are the first to show that \ac{vjepa} performs on-par with other \ac{fer} methods.
    \end{enumerate}
    This paper is structured as follows:
    In \Cref{sec:related literature} we give a short overview of other self-supervised methods for \ac{fer} exemplifying the widespread use of pixel-based reconstruction techniques during pre-training in this field.
    Additionally, we give a short introduction into \acp{jepa}.
    In \Cref{sec:methodology} we describe \acp{vjepa}, attentive probing (a feature pooling and classification scheme) and how to classify full videos from a set of clips containing $16$ frames each.
    Afterwards, in \Cref{sec:experiments} we describe all of our experiments, the datasets that were used for them and the metrics we applied.
    Further, in \Cref{sec:results and discussion} we describe and analyze our results.
    Finally, in \Cref{sec:conclusion} we conclude this paper.
\section{Related Literature}\label{sec:related literature}\noindent
    In the following section we give a brief overview of different \ac{fer} methods that employ self-supervised pre-training.
    The given examples further exemplify how pixel-based reconstruction pretext tasks are used for \ac{fer}, and motivate an introduction into \acp{jepa}.
\subsection{Self-supervised FER methods}\label{ssec:self-supervisedfermethods}\noindent
    SVFAP~\cite{sun2024svfap} is a video encoder model for a handful of different affect-related computer vision tasks.
    The authors propose a specialized Transformer architecture they call the Temporal Pyramid and Spatial Bottleneck Transformer (TPSBT).
    They use this model architecture as their video encoder.
    They borrow their pretext tasks from VideoMAE~\cite{tong2022videomae}, i.e. reconstructing tube-masked regions of given the embeddings of unmasked tokens.
    
    \cite{chen2023self} use a three-stage framework for~\acs{fer}.
    The authors define a set of four pretext tasks (denoising, rotation prediction, jigsaw puzzling, masked region reconstruction) for pre-training a~\ac{vit} type model as a vision encoder in a self-supervised manner.
    After pre-training, they finetune their models on a laboratory-controlled~\ac{fer} dataset.
    To achieve~\ac{fer} in the wild, they extract spatio-temporal embedding from a support set of each class.
    Then, they calculate class descriptors from those embeddings and perform few-shot classification by classifying samples as the class with the least distant class descriptor.
    
    \acs{maedfer}~\cite{sun2023mae} is a methodology for~\acl{dfer} using a self-supervised pre-training scheme.
    It is another method taking inspiration from \acs{videomae}~\cite{tong2022videomae}.
    It extends this approach by a more efficient Transformer architecture, LGI-Former, reducing computational cost by approx. $38\%$.
    To this end, they introduce tokens that represent local spatio-temporal regions and only calculate self-attention between those representations.
    They show that their model outperforms other state-of-the-art methods on six datasets by a significant margin, while staying comparable to performance achieved using a fully-fledged \acs{videomae}.
    
    All the methods mentioned above (and more~\cite{sun2024hicmae, xiang2024multimae}), all of which employ self-supervised pretraining, heavily rely on pixel-level reconstructions of an input signal during their pre-training stages.
    We argue that pixel-level reconstructions during pretraining are not necessary to produce a well-performing model for \ac{fer} tasks.
\subsection{Joint-Embedding Predictive Architectures}\label{ssec:joinembeddingpredictivearchitectures}\noindent
    \begin{figure}
       \centerline{
\newcommand{\fixedWidth}{2cm}
\newcommand{\fixedHeight}{1cm}

\tikzset{
    direct/.style={
        rectangle,
        minimum height=\fixedHeight,
        minimum width=\fixedWidth,
        rounded corners=1.5mm,
        align=center,
        draw,
        thick
    }
}

    \begin{tikzpicture}

        \node[draw, color=darkgray, circle] (x) at (0, -1.5)  {$x$};

        \node[draw, color=darkgray, circle] (y) at (4, -1.5)  {$y$};

        \node[draw, color=darkgray, fill=pink, circle] (z) at (-2, 2)  {$z$};

        \node[direct, color=darkgray, fill=white] (encoder1) at (0, 0) {$E_{\theta_x}$}; 

        \node[direct, color=darkgray, fill=white] (encoder2) at (4, 0) {$\hat{E}_{\theta_y}$}; 

        \node[direct, color=darkgray, fill=white] (predictor) at (0, 2) {$P_\phi$}; 

        \node[direct, color=darkgray, fill=pink] (loss) at (4, 2) {$\mathcal{L}(\hat{s}_y, s_y)$}; 

        \draw[->, color=darkgray] (x.north) -- (encoder1.south);

        \draw[->, color=darkgray] (y.north) -- (encoder2.south);

        \draw[->, color=red] (z.east) -- (predictor.west);
 
        \draw[->, color=darkgray] (encoder1.north) -- (predictor.south) node[midway, left] {$s_x$};

        \draw[->, dashed, color=darkgray] (encoder2.north) -- (loss.south) node[midway, left] {$s_y$};

        \draw[->, dashed, color=darkgray] (predictor.east) -- (loss.west) node[midway, above] {$\hat{s}_y$};

    \end{tikzpicture}
       }
       \caption{
          General overview of a \acl{jepa}.
       }\label{fig:jepa}
    \end{figure}
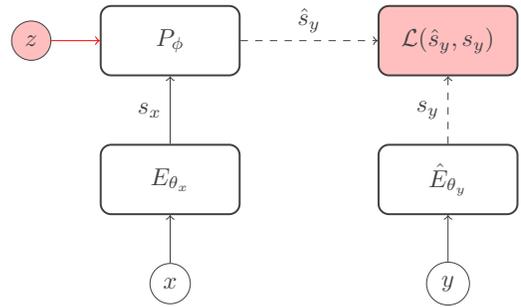
    \acfp{jepa} are a type of model architecture and training regimen proposed by LeCun in~\cite{lecunPathAutonomousMachine}.
    The author makes an argument for training encoders $E_{\theta_x}$ and $\hat{E}_{\theta_y}$ of different input signals $x$ and $y$ jointly with a predictor $P_\phi$.
    The predictor predicts the latent representation $\hat{s}_y$ of an input $y$ from the latent representation $s_x$ of an input $x$ given a conditioning variable $z$.
    A general overview of this concept is shown in~\Cref{fig:jepa}.
    The trained encoder $E_{\theta_x}$ promises to exclusively capture information contained in the input signal $x$ that is predictive of the latent representation of $y$. 
    First experiments on image and video encoding show that training encoders this way produces encoders capable of reaching state-of-the-art performance on downstream image classification and action recognition tasks that are more data efficient than other methods~\cite{assran2023selfsupervisedlearningimagesjointembedding,bardes2024revisitingfeaturepredictionlearning}.
    
    \acfp{vjepa} proposed by~\cite{bardes2024revisitingfeaturepredictionlearning} are \acp{jepa} trained on video.
    They are trained to encode and predict video features of masked regions of video within latent space.
    In this work, we use the encoder $E_{\theta_y}$ taken from a pre-trained \ac{vjepa}, and train a set of shallow classifiers for \ac{fer} on top of it.
    We show that pixel-level reconstruction of video is not needed as a pretext task for \ac{fer}.
\section{Methodology}\label{sec:methodology}\noindent
    In this section, to facilitate a better understanding of the video encoder, we first go into more detail of \ac{vjepa} pre-training.
    We then explain how classifiers were trained on top of the video encoder for the \ac{fer} task.
    Furthermore, we explain how we perform video classification using clips sampled from full videos.
\subsection{\aclp{vjepa}}\label{ssec:vjepa}\noindent
   \begin{figure}
      \centerline{
\newcommand{\fixedWidth}{1cm}
\newcommand{\fixedHeight}{1cm}

\tikzset{
    direct/.style={
        rectangle,
        minimum height=\fixedHeight,
        minimum width=\fixedWidth,
        rounded corners=1.5mm,
        align=center,
        draw,
        thick
    }
}
    \begin{tikzpicture}

        \node[
            rectangle, 
            text width=2cm, 
            minimum height=1cm, 
            minimum width=2cm, 
            rounded corners=1.5mm, 
            align=center, 
            draw, 
            thick, 
            color=darkgray, 
            fill=white
        ] (masking) at (0, 0) {Remove masked tokens};

        \node[direct, color=darkgray, fill=white] (encoder) at ($(masking.center) + (0, 1.7)$) {$E_\theta$};

        \node[direct, color=darkgray, fill=white] (predictor) at ($(encoder.center) + (0, 1.7)$) {$P_\phi$};

        \node[direct, color=darkgray, fill=pink] (loss) at ($(predictor.center) + (0, 1.7)$) {$L_1(\hat{s}_y, s_y)$};

        \node[
            rectangle, 
            text width=2cm, 
            minimum height=1cm, 
            minimum width=2cm, 
            rounded corners=1.5mm, 
            align=center, 
            draw, 
            thick, 
            color=darkgray, 
            fill=white
        ] (un_masking) at ($(loss.center) + (0, 1.7)$) {Remove unmasked tokens};

        \node[direct, color=darkgray, fill=white!80!blue] (target_encoder) at ($(un_masking.center) + (0, 1.7)$) {$E_{\hat{\theta}}$};

        \node[draw, color=darkgray, circle] (x) at ($(loss.center) + (-2, 0)$) {$x$};

        \node[draw, color=darkgray, circle] (m) at ($(loss.center) + (2, 0)$) {$m$};

        \draw[->, color=darkgray] (x.south) -- (x.south |- masking.west) -- (masking.west);
        \draw[->, color=darkgray] (x.north) -- (x.north |- target_encoder.west) -- (target_encoder.west);
        \draw[->, dashed, color=darkgray] (m.south) -- (m.south |- masking.east) -- (masking.east);
        \draw[->, dashed, color=darkgray] (m.north) -- (m.north |- un_masking.east) -- (un_masking.east);
        \draw[->, color=darkgray] (target_encoder.south) -- (un_masking.north);
        \draw[->, dashed, color=darkgray] (un_masking.south) -- (loss.north) node[midway, left] {$s_y$};
        \draw[->, dashed, color=darkgray] (predictor.north) -- (loss.south) node[midway, left] {$\hat{s}_y$};
        \draw[->, color=darkgray] (masking.north) -- (encoder.south);
        \draw[->, color=darkgray] (encoder.north) -- (predictor.south) node[midway, left] {$s_x$};


        \draw[-, dashed, color=darkgray] ($(target_encoder.center -| m.center) + (1, 0)$) -- ($(masking.center -| m.center) + (1, 0)$);

        \node[direct, color=darkgray, fill=white!80!blue] (p_target_encoder) at ($(m.center) + (3.5, -3)$) {$E_{\hat{\theta}}$};

        \node[draw, color=darkgray, circle] (p_x) at ($(p_target_encoder.center) + (0, -1.7)$) {$x$};

        \node[draw, color=darkgray, circle] (p_q) at ($(p_x.center) + (-1.5, 0)$) {$q$};
        
        \node[
            rectangle, 
            text width=2cm, 
            minimum height=1cm, 
            minimum width=2cm, 
            rounded corners=1.5mm, 
            align=center, 
            draw, 
            thick, 
            color=darkgray, 
            fill=white
        ] (p_cross_attention) at ($(p_target_encoder.center) + (0, 1.7)$) {Cross Attention};

        \node[
            rectangle, 
            text width=2cm, 
            minimum height=1cm, 
            minimum width=2cm, 
            rounded corners=1.5mm, 
            align=center, 
            draw, 
            thick, 
            color=darkgray, 
            fill=white
        ] (p_mlp) at ($(p_cross_attention.center) + (0, 1.7)$) {MLP};

        \node[
            rectangle, 
            text width=2cm, 
            minimum height=1cm, 
            minimum width=2cm, 
            rounded corners=1.5mm, 
            align=center, 
            draw, 
            thick, 
            color=darkgray, 
            fill=white
        ] (p_prediction) at ($(p_mlp.center) + (0, 1.7)$) {Prediction};

        \draw[->, color=darkgray] (p_q.north) -- (p_q.north |- p_cross_attention.west) -- (p_cross_attention.west);

        \draw[->, color=darkgray] (p_x.north) -- (p_target_encoder.south);

        \draw[->, color=darkgray] (p_target_encoder.north) -- (p_cross_attention.south) node[midway, left] {$s_x$};;

        \draw[->, color=darkgray] (p_cross_attention.north) -- (p_mlp.south);

        \draw[->, color=darkgray] (p_mlp.north) -- (p_prediction.south);

    \end{tikzpicture}
      }
      \caption{
         Training Setup for \acp{vjepa} (left) and Attentive Probing (right)
      }\label{fig:vjepa}
   \end{figure}
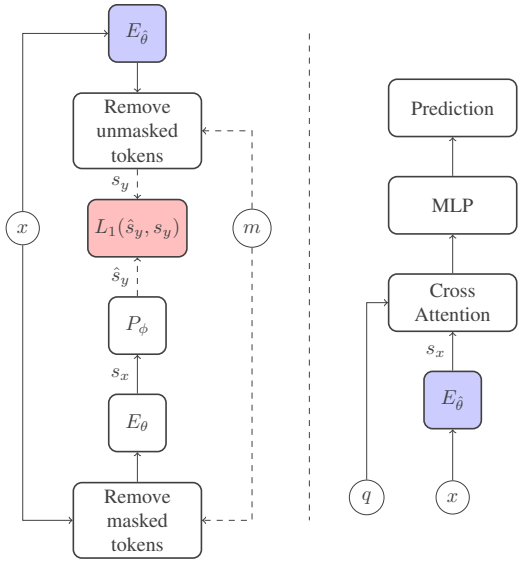
   As a video encoder we use a \acf{vjepa}.
   \acp{vjepa} are particular types of \acp{jepa} trained for video encoding.
   They are trained by predicting the representations $s_y$ of tube-masked regions of video using mask $m$ from the representations $s_x$ of the remaining video $x$.
   We show this setup on the left side of \Cref{fig:vjepa}.
   Both predictor $P_\phi$ and encoder models $E_\theta$ and $E_{\hat{\theta}}$ are (vision) transformer architectures~\cite{dosovitskiy2020image,vaswani2017attention}.
   In fact, the target encoder $E_{\hat{\theta}}$, shown in blue in \Cref{fig:vjepa}, has the same architecture as input encoder $E_{\theta}$ and its weights are bootstrapped to $E_\theta$ via an exponential moving average~\cite{tarvainen2017mean}.
   During pretraining, masking tokens, positionally encoded to describe the masked regions of video, are used as the conditioning variable $z$ of the predictor.
   No gradients are propagated backwards through $\hat{\theta}$.
   The input encoder $E_\theta$ is only allowed to look at the already masked input video, while the target encoder $E_{\hat{\theta}}$ is allowed to look at the whole video and the output representations are then masked and fed into a $L_1$ loss function.
   Videos are subsampled from longer videos as clips of equal length to allow for more efficient training.
   This setup is aimed at forcing the input encoder to produce embeddings of the unmasked regions of video that are predictive of the embeddings of the masked regions of video.
\subsection{Attentive Probing}\label{ssec:attentiveprobing}\noindent
   To perform classification, instead of applying a linear operation like global average pooling followed by a set of linear layers, we use an attentive probe~\cite{bardes2024revisitingfeaturepredictionlearning}.
   An attentive probe consists of an attentive pooling layer, pooling features $s_x$ into a single output token using a single learnable query token $q$.
   This replaces widely used linear pooling operations like global average pooling.
   The single output token is then fed to a \ac{mlp} for the classification task.
   Using an attentive probe has been shown to achieve significantly better results ($~+17\%$ on K400) for \ac{vjepa} type models~\cite{bardes2024revisitingfeaturepredictionlearning}.
   The authors hypothesize that this is because there is no a priori reason for the encoder of \ac{vjepa} type models to learn linearly separable feature embeddings~\cite{chen2020simple}.
   Thus, applying a linear pooling operation to such embeddings would not yield optimal results.
   We show this approach on the right side of \Cref{fig:vjepa}.
\subsection{Classifying Whole Videos}\label{ssec:classifying whole videos}\noindent
   To achieve results that are comparable to other \ac{fer} approaches during validation, we need to classify whole videos.
   To classify whole videos, we evaluate Maximum Voting (MV) and Posteriors-based Voting (PBV).
   As in~\cite{bardes2024revisitingfeaturepredictionlearning}, videos are split into overlapping clips consisting of $16$ frames each.
   Clips are sampled with an inter-frame skip of $4$, resulting in clips of approximately $3$ seconds in length (at $24$ frames per second).
   Videos that are too short to contain a single clip are padded with their last frame being repeated.
   For the MV strategy, we predict the class of each clip, and then predict the class with the highest accumulated score for the whole video.
   For the PBV strategy, we add up the predicted class probabilities of all clips from a given video, and only then predict a given class.
\section{Experiments}\label{sec:experiments}\noindent
   We train a set of five classifiers per dataset using $5$-fold subject-independent cross-validation, one classifier for each fold.
   We use two lab-controlled datasets for training and cross-validation.
   This results in a total of ten classifiers being trained.
   For each, we evaluate both MV and PBV strategies.
   Additionally, we perform cross-dataset validation, using the dataset a given model was \textit{not} trained on as an additional validation fold.
   This way we can get an even better predictor of performance on previously unseen individuals and try to eliminate any dataset bias.

   We do not perform ablation experiments on the impact of using attentive probes vs. linear pooling operations.
   We do not do this, because~\cite{bardes2024revisitingfeaturepredictionlearning} showed performance increases of approximately $17\%$ on the K400~\cite{kay2017kinetics} and SSv2~\cite{goyal2017something} datasets.
   We argue that these performance boosts are large enough to justify this.
   For more information on the impact of attentive probing not only on \ac{vjepa} type models, we'd like to refer readers to Section 12.1 of~\cite{bardes2024revisitingfeaturepredictionlearning}.
\subsection{Training Setup}\label{ssec:model configuration}\noindent
   For all of our experiments we use an already pre-trained \ac{vjepa} video encoder of the ViTH model type\footnote{Available at \href{https://github.com/facebookresearch/jepa}{https://github.com/facebookresearch/jepa}}.
   It was trained on a set of approx. $2$ million videos taken from the HowTo100M~\cite{miech2019howto100m}, Kinetics-400/600/700~\cite{kay2017kinetics} and Something-Something-v2~\cite{goyal2017something} datasets~\cite{bardes2024revisitingfeaturepredictionlearning}.
   These datasets, in particular HowTo100M, include large amounts of videos containing faces, enhancing the models' ability to create latent representations of them. 
   Good latent representations of faces are key to \ac{fer}~\cite{sun2023mae, sun2024hicmae}.
   We keep the weights of the \ac{vjepa} type video encoder frozen the whole time, only backpropagating gradients through the classification head.
   In line with the pre-training, videos are subsampled into clips of $16$ frames with an inter-frame skip of $4$.
   Videos that are too short to contain a single clip are padded with their last frame being repeated.
   Clips are normalized, randomly cropped and then resized into $224\times 224$ pixels.
   They are then divided into tokens of size $16\times 16\times 2$, i.e. each input token spans two frames of video.
   Tokens are positionally encoded using a three-dimensional sinusoidal positional embedding.
   After feeding the clips to the video encoder model, the output embeddings are used for classification.
   As the classification head we use an attentive probe with an \ac{mlp} of depth $3$.
   During the training of the classification heads, we randomly select a total of $8$ clips from each video per epoch.
   Since some videos from both datasets are too short to sample $8$ non-overlapping clips, we allow clips to overlap.
   Randomly selecting only $8$ clips from each video also means that we discard parts of some videos during each epoch.
   We do this to not introduce any additional bias towards classes that are represented by longer videos.
   We train our classifiers for a total of $20$ epochs.
   During validation we do not only consider randomly selected clips from each video, but instead take into account every (overlapping) possible clip from a given video.
\subsection{Data}\label{ssec:data}\noindent
\begin{figure*}[h]
    \centering
    \begin{subfigure}{\columnwidth}
        \centering
        \includegraphics[width=.9\linewidth]{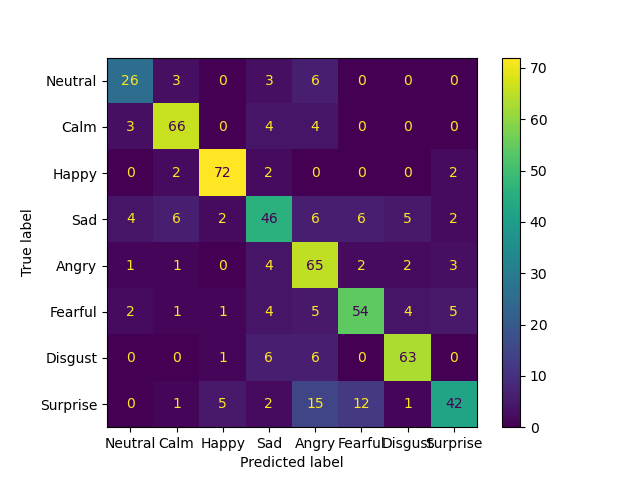}
        \caption{\ac{ravdess}.}
        \label{fig:appendix:ravdess}
    \end{subfigure}%
    \begin{subfigure}{\columnwidth}
        \centering
        \includegraphics[width=.9\linewidth]{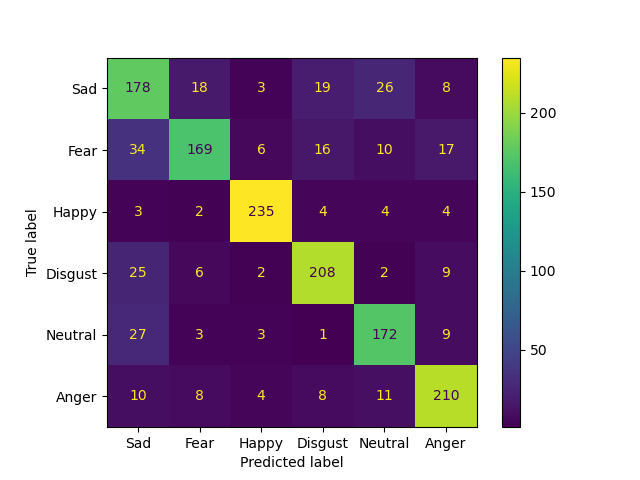}
        \caption{\ac{cremad}}
        \label{fig:appendix:cremad}
    \end{subfigure}
    \caption{Confusion matrices of same-dataset experiments.}
    \label{fig:confusionmatrices}
\end{figure*}
   The \textbf{\acf{ravdess}}~\cite{livingston2018ravdess} is a multi-modal database containing a total of $7356$ recordings of $24$ actors vocalizing lexically-matched statements in neutral North American accent in speech and song.
   In this work, to be able to achieve comparable results during our cross-dataset evaluation experiments, only the recordings of speech were used, a total of $2880$ recordings.
   They are labeled with $8$ emotional states (calm, happy, sad, angry,  fearful, surprise, disgust, and neutral).
   Each expression is produced at two levels of intensity by each actor, except the neutral expression, which is only produced once.

   The \textbf{\acf{cremad}}~\cite{cao2014crema} is an audio-visual dataset containing a total of $7442$ recordings of $91$ actors vocalizing a set of $12$ sentences labeled with a total of $6$ basic emotional states (happy, sad, anger, fear, disgust, and neutral).
   The labels were validated using a total of $2443$ raters.

   To create training and validation folds for the $5$-fold cross-validation, the data needs to be split subject-independently.
   Splitting the data independent of subject is important as validation should test how well models will generalize to new individuals.
   For \ac{ravdess} we use the splits provided by~\cite{lunajimenez2021multimodelemotionrecognition}, because they already fit this criterion.
   \begin{table}[t]
      \caption{
         Subject IDs in each 5-fold cross validation split for CREMA-D.
      }
      \begin{center}
         \begin{tabularx}{\columnwidth}{cX}
            \toprule
            \textbf{Split} & \textbf{Subject IDs}\\
            \midrule
            1 & 2, 3, 8, 10, 14, 16, 22, 32, 51, 54, 60, 66, 68, 70, 72, 77, 78, 86, 88\\
            2 & 4, 11, 12, 13, 19, 30, 33, 35, 39, 40, 48, 49, 53, 57, 67, 71, 73, 81\\
            3 & 5, 6, 17, 26, 28, 34, 42, 44, 50, 59, 64, 69, 82, 83, 84, 87, 90, 91\\
            4 & 7, 9, 15, 21, 23, 29, 36, 37, 38, 41, 45, 52, 55, 58, 61, 62, 74, 76\\
            5 & 1, 18, 20, 24, 25, 27, 31, 43, 46, 47, 56, 63, 65, 75, 79, 80, 85, 89\\
            \bottomrule
         \end{tabularx}
         \label{tab:subjects crema-d}
      \end{center}
   \end{table}
   For \ac{cremad} we make $5$ pseudo-random, subject-independent splits such that each training fold roughly follows the same distribution of parameters for size of fold, age, sex, race, and ethnicity.
   This does not mean that each validation fold necessarily follows the same distribution across all of these parameters.
   A discussion of this can be found in \Cref{ssec:ravdess and crema-d} and our Ethical Impact Statement.
   The subject IDs for each split are provided in~\Cref{tab:subjects crema-d}.
   For brevity subject IDs are given with all leading, non-identifying digits removed.
\subsection{Metrics}\label{ssec:metrics}\noindent
We report \ac{uar} and \ac{war}, as they are commonly used for \ac{fer}.
Given classes $C$, class-wise true positives $TP(c, \hat{D}, D)$, and false negatives $FN(c, \hat{D}, D)$ for an arbitrary class $c\in N$, predictions $\hat{D}$, and ground-truths $D$, \ac{uar} is given by:
\begin{equation}
  UAR = \frac{1}{|C|}\sum_{c\in C} \frac{TP(c, \hat{D}, D)}{TP(c, \hat{D}, D) + FN(c, \hat{D}, D)}\label{eq:uar}
\end{equation}
Since there are slight class imbalances in \ac{ravdess} and \ac{cremad} and \ac{uar} does not account for them, we also report \ac{war}.
Given the number of occurrences $|D_c|$ of class $c\in C$ in $D$, \ac{war} is given by:
\begin{equation}
  WAR = \sum_{c\in C} \frac{|D_c|}{|D|}\cdot\frac{TP(c, \hat{D}, D)}{TP(c, \hat{D}, D) + FN(c, \hat{D}, D)}\label{eq:war}
\end{equation} 
All metrics are always averaged across all five validation folds.
\section{Results and Discussion}\label{sec:results and discussion}\noindent
In the following section we perform the described experiments on both datasets and evaluate the MV and PBV strategies in \Cref{ssec:ravdess and crema-d}.
We then perform cross-dataset evaluation in \Cref{ssec:cross-dataset evaluation}.
Finally, in \Cref{ssec:discussion and comparison} we discuss our results and compare them to other state-of-the-art self-supervised \ac{fer} methods.
\subsection{RAVDESS and CREMA-D}\label{ssec:ravdess and crema-d}\noindent
\begin{table}[t]
   \caption{Validation Results on \ac{ravdess} and \ac{cremad}}
   \begin{center}
      \begin{tabular}{cccc}
         \toprule
         \textbf{Dataset} & \textbf{Voting Strategy} & \textbf{\ac{uar}}$\uparrow$ & \textbf{\ac{war}}$\uparrow$\\
         \midrule
         \ac{ravdess} & PBV & $76.40$ & $72.93$\\
                      & MV & $76.38$ & $73.80$ \\
         \midrule
         \ac{cremad} & PBV & $79.39$ & $78.86$\\
                     & MV & $79.13$ & $78.47$ \\
         \bottomrule
      \end{tabular}
      \label{tab:evaluation results}
   \end{center}
\end{table}
\Cref{tab:evaluation results} shows the validation metrics retrieved by training the attentive probes using \ac{ravdess}.
Using the posteriors-based voting strategy we achieve $76.40$ and $72.93$ \ac{uar} and \ac{war} respectively.
Following a maximum voting (MV) strategy we achieve $76.38$ and $73.80$.
\Cref{tab:evaluation results} also shows  the validation metrics retrieved by training the attentive probes using \ac{cremad}.
Using posteriors-based voting (PBV) we achieve $79.39$ and $78.86$ \ac{uar} and \ac{war} respectively.
Following a maximum voting (MV) strategy we achieve $79.13$ and $78.47$ \ac{uar} and \ac{war} respectively.

These results show that there is no major advantage in using one strategy over the other.
Therefore, in the following experiment we only report the results using the PBV strategy.
For more detailed information on how models performed we also add the average confusion matrices using the PBV strategy in~\Cref{fig:confusionmatrices}.
\subsection{Cross-Dataset Evaluation}\label{ssec:cross-dataset evaluation}\noindent
\begin{table*}[t]
   \caption{
      Comparison to other self-supervised \ac{fer} methods.
      Best overall methods are \underline{underlined}, best methods using the video modality are presented in \textbf{bold}.
   }
   \begin{center}
      \begin{tabular}{cccccccc}
         \toprule
         \multicolumn{4}{c}{\textbf{\acs{ravdess}}} & \multicolumn{4}{c}{\textbf{\acs{cremad}}}\\
         \multicolumn{4}{c}{
            \begin{tabular}{cccc}
               \toprule
               \textbf{Method} & \textbf{Modality} & \textbf{\acs{uar}}$\uparrow$ & \textbf{\acs{war}}$\uparrow$\\
               \midrule
               \textit{Human Perception}~\cite{livingston2018ravdess} & Video + Audio & $77.88$ & $77.94$ \\
               MultiMAE-DER~\cite{xiang2024multimae} & Video + Audio & $82.23$ & $83.61$\\
               HiCMAE-B~\cite{sun2024hicmae} & Video + Audio & \underline{$87.96$} & \underline{$87.99$} \\
               \midrule
               \midrule
               MultiMAE-DER~\cite{xiang2024multimae} & Video & -- & $74.13$\\
               HiCMAE-B~\cite{sun2024hicmae} & Video & $71.35$ & $70.97$\\
               SVFAP-B~\cite{sun2024svfap} & Video & $75.15$ & $75.01$\\
               \acs{maedfer}~\cite{sun2023mae} & Video & $75.91$ & \boldmath{$75.56$}\\
               \bottomrule
               Ours & Video & \boldmath{$76.40$} & $72.93$\\
            \end{tabular}
         }
         &
         \multicolumn{4}{c}{
            \begin{tabular}{cccc}
               \toprule
               \textbf{Method} & \textbf{Modality} & \textbf{\acs{uar}}$\uparrow$ & \textbf{\acs{war}}$\uparrow$\\
               \midrule
               \\
               MultiMAE-DER~\cite{xiang2024multimae} & Video + Audio & $79.12$ & $79.36$\\
               HiCMAE-B~\cite{sun2024hicmae} & Video + Audio & \underline{$84.91$} & \underline{$84.89$} \\
               \midrule
               \midrule
               MultiMAE-DER~\cite{xiang2024multimae} & Video & -- & $77.83$\\
               HiCMAE-B~\cite{sun2024hicmae} & Video & $77.25$ & $77.21$\\
               SVFAP-B~\cite{sun2024svfap} & Video & $77.31$ & $77.37$\\
               \acs{maedfer}~\cite{sun2023mae} & Video & $77.33$ & $77.38$ \\
               \bottomrule
               Ours & Video & \boldmath{$79.39$} & \boldmath{$78.86$}\\
            \end{tabular}
         }\\
         \bottomrule
      \end{tabular}\label{tab:comparisons} 
   \end{center}
\end{table*}
\begin{figure*}
    \centering
    \begin{subfigure}{\columnwidth}
        \centering
        \includegraphics[width=.9\linewidth]{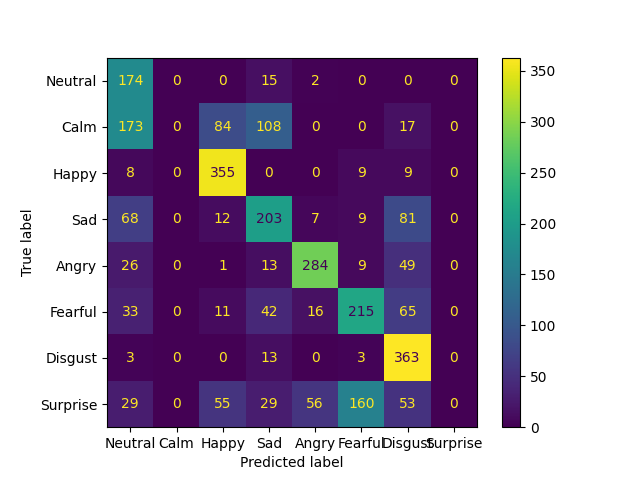}
        \caption{\ac{ravdess}}
        \label{fig:appendix:ravdess-cross}
    \end{subfigure}%
    \begin{subfigure}{\columnwidth}
        \centering
        \includegraphics[width=.9\linewidth]{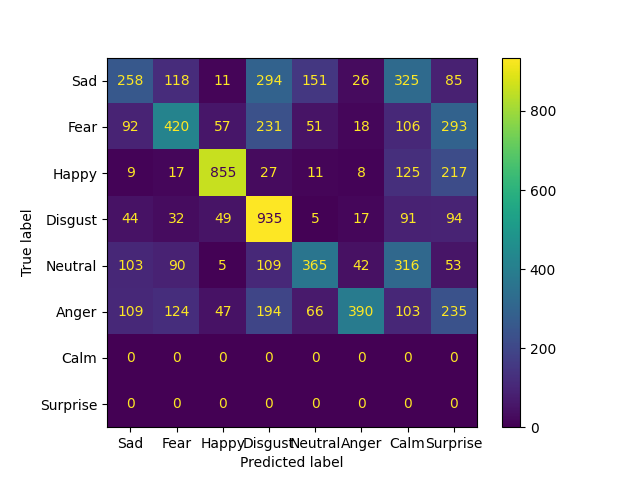}
        \caption{\ac{cremad}}
        \label{fig:appendix:cremad-cross}
    \end{subfigure}
    \caption{Average confusion matrixes of cross-dataset evaluations.}
    \label{fig:confusionmatrices-cross}
\end{figure*}
\begin{table}
   \caption{Cross-Dataset Evaluation Results.}
   \begin{center}
      \begin{tabular}{ccc}
         \toprule
         \textbf{Test Dataset} & \textbf{\ac{uar}}$\uparrow$ & \textbf{\ac{war}}$\uparrow$ \\
         \midrule
         CREMA-D & $61.69$ & $59.82$\\
         *       & $58.72$ & $54.90$\\
         \midrule
         RAVDESS & $77.82$ & $75.59$\\
         *       & $73.92$ & $70.92$
      \end{tabular}
      \label{tab:cross-dataset test results}
   \end{center}
\end{table}
To be able to perform cross-dataset validation between \ac{ravdess} and \ac{cremad} some predictions and labels need to be specifically handled.
In particular, \ac{cremad} does not have the ``calm'' and ``surprise'' labels that \ac{ravdess} uses.

To handle the ``surprise'' label, we simply ignore all predictions made for this class by classifiers trained on \ac{ravdess}.
Similarly, during evaluation of classifiers that were trained on \ac{cremad}, we do not include predictions in our results when videos were labeled with the ``surprise'' label. 
We argue that this is valid for cross-dataset validation, because we assume what was stated by the authors of~\cite{cao2014crema} to be true for \ac{ravdess} too; I.e. that ``surprise'' can not be considered to be a sufficiently specific acting direction, as it could relate to any of the other emotions with rapid onset.

To handle the ``calm'' label we follow two different approaches.
In addition to following the same approach as for ``surprise'', we also test combining ``calm'' predictions and labels into one class with ``neutral''.
We justify this by a fact about \ac{ravdess}.
The ``calm'' label was introduced to provide a second baseline emotion, that does not convey a mild negative valence, but is perceptually identical to ``neutral''~\cite{livingston2018ravdess}.
Thus, we can combine both baseline emotions into one to test how well each classifier recognizes them, regardless of valence.

The results are shown in \Cref{tab:cross-dataset test results}.
Results for which baseline emotions were combined into one are marked with ``*''.
Surprisingly, models that were trained on \ac{cremad} achieve similar performance on \ac{ravdess} ($75.59$~\ac{war}) to models that were trained on \ac{ravdess} itself~($72.93$~\ac{war}).
This is not the case the other way around.
Models that were trained on \ac{ravdess} perform much worse when evaluated on \ac{cremad} ($-19.04$ decrease in \ac{war}).
It is apparent that performance decreases significantly when baseline emotion predictions and labels are combined into one class, versus when samples and predictions are dropped (approx. $-4.5$ decrease in \ac{war} in both cases).
\begin{figure*}
    \centering
    \begin{subfigure}{\columnwidth}
        \centering
        \includegraphics[width=.9\linewidth]{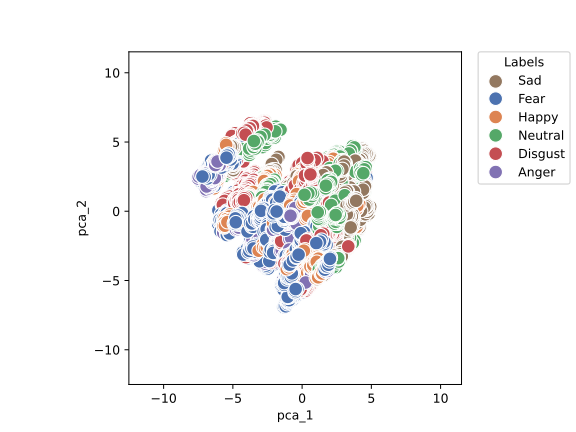}
        \caption{With global average pooling}
        \label{fig:appendix:gax-vis}
    \end{subfigure}%
    \begin{subfigure}{\columnwidth}
        \centering
        \includegraphics[width=.9\linewidth]{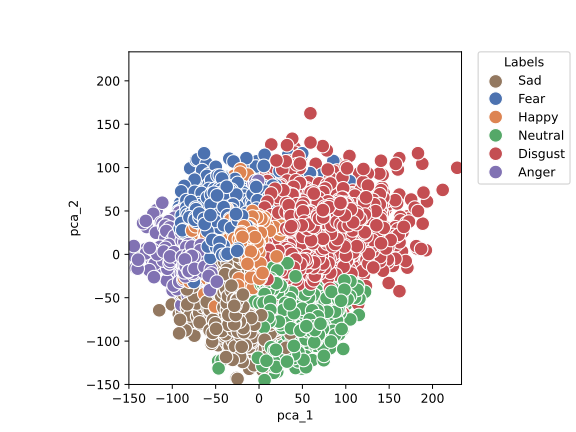}
        \caption{With attentive probing}
        \label{fig:appendix:attp-vis}
    \end{subfigure}
    \caption{Visualization of the first two principal components (PCA) using global average pooling (left), and attentive probing (right). Embeddings were produced using the best model trained on CREMA-D during k-fold validation on the respective validation dataset.}
    \label{fig:appendix:vis}
\end{figure*}
\subsection{Comparison and Discussion}\label{ssec:discussion and comparison}\noindent
For our comparison, we use a set of other methods for \ac{fer} that were trained using pixel-level reconstruction pretext tasks in a self-supervised manner and were evaluated on \ac{ravdess} and \ac{cremad}.
We divide our comparison into methods that use audio and video as well as methods like ours that use video only.
We show this comparison in \Cref{tab:comparisons}.

We note that we achieve good state-of-the-art results, in the case of \ac{cremad} outperforming all other purely vision-based methods we compare ourselves to.
For both datasets the highest per-class recall was achieved for ``happy'' ($87.58$ and $92.84$ for \ac{ravdess} and \ac{cremad} respectively).
The lowest per-class recall was achieved for the ``sad'' emotions ($63.24$ and $64.22$).
These results are in line with other methods.
For more information about per-class performance, we refer the reader to~\Cref{fig:appendix:ravdess} and~\Cref{fig:appendix:cremad},  where we add confusion matrices averaged across all training runs.

The results achieved during cross-dataset evaluation have to be taken with a grain of salt, as some predictions are dropped during evaluation, essentially throwing out incorrect predictions, that would otherwise decrease \ac{uar} and \ac{war} scores.
We argue that our results are still a good indicator of our models' generalization capabilities.
We observe that our models trained on \ac{cremad} confused $47.39\%$ of the videos labeled as ``calm'' as ``neutral'', validating our assumption that those emotions are indeed very similar to models that do not know the ``calm'' label. 
Additionally, we observe that those models predict $41.88\%$ of videos labeled as ``surprise'' as ``fear'', predictions which were thrown out for both evalutations with and without ``calm'' and ``neutral'' emotions combined.
For the models trained on \ac{ravdess} something similar can be observed, i.e. that almost $30\%$ of videos labeled as ``fear'' were predicted as ``surprise''.
For more detail, we add the confusion matrices averaged across all cross-dataset evaluations in~\Cref{fig:appendix:ravdess-cross} and~\Cref{fig:appendix:cremad-cross}.

To add to the comparability of our work, we add visualizations of the first two principal components of feature embeddings of all clips in the validation dataset of the first data split of \ac{cremad} (see~\Cref{fig:appendix:vis}).
We do not use t-SNE visualizations, as they are highly dependent on their hyperparameters.

Overall we interpret our results to show, that \acp{vjepa} are good vision encoders for the \ac{fer} task, which was not trained on pixel-based pretext tasks.
However, it has to be noted that these are results solely obtained on lab-controlled datasets.
Whether \acp{vjepa} are also good video encoders for \ac{fer} in-the-wild remains to be seen.
\section{Conclusion}\label{sec:conclusion}\noindent
In this work we showed that pixel-based reconstructions are not a necessary pretext task for \ac{fer}.
To this end, we used an already pre-trained \ac{vjepa} model as our video encoder.
We achieved state-of-the-art performance by using an attentive probe to train shallow classifiers on the \ac{ravdess} and \ac{cremad} datasets.
Additionally, we applied the trained models in a cross-dataset evaluation.
We discussed our results and compared them to other state-of-the-art \ac{fer} methods using pixel-based pretext tasks during their pre-training.
\section*{Ethical Impact Statement}\label{sec:ethicalimpactstatement}\noindent
%
\begin{figure}[h]
    \centering
    \includesvg[width=.9\columnwidth]{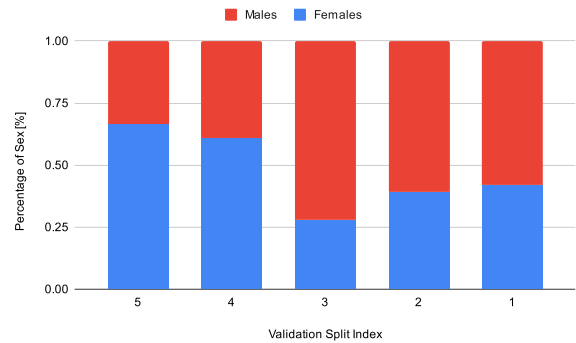}
    \caption{Sex distribution of \ac{cremad} data splits.}
    \label{fig:appendix:sex}
\end{figure}
%
\noindent
There were two main ethical concerns we considered when developing this system for \ac{fer}.
Firstly, we want to point out that experiments were conducted solely on laboratory-controlled datasets.
Therefore, generalizability could be low  for real world application scenarios, whether \acp{vjepa} are good video encoders for in-the-wild \ac{fer} tasks remains to be tested.

Secondly, as was pointed out in \Cref{ssec:data}, validation splits for \ac{cremad} were not balanced in their distribution of (known) parameters.
For example, there is a strong underrepresentation of female actors in split $3$.
We show this in~\Cref{fig:appendix:sex}.
However, this does not have a major impact on model performance, as there was a standard deviation in \ac{war} of approx. $2.3$ for both datasets.
This shows that the video encoder and trained classifiers are indifferent to changes in sex, ethnicity, race, and age.
However, it should be noted that both datasets were recorded using North American persons.
This could have an impact on generalizability when applied to persons of other nationality/culture.

\bibliographystyle{IEEEtran}
\bibliography{literature}

\end{document}